\documentclass[runningheads]{llncs}

\usepackage{eccv}

\usepackage{eccvabbrv}

\usepackage{graphicx}
\usepackage{booktabs}
\usepackage{caption}
\usepackage{tcolorbox}
\usepackage{xcolor}

\usepackage[accsupp]{axessibility}  

\usepackage[breaklinks,colorlinks,citecolor=eccvblue]{hyperref}

\begin{document}

\title{IUP-Pose: Decoupled Iterative Uncertainty Propagation for Real-time Relative Pose Regression via Implicit Dense Alignment} 

\titlerunning{IUP-Pose}

\author{Jun Wang\thanks{arXiv preprint, v1.}\inst{1} \and Xiaoyan Huang\inst{2}}

\authorrunning{J.~Wang and X.~Huang}

\institute{
\email{1814409917@qq.com}
\and
\email{20193712012@m.scnu.edu.cn}
}

\maketitle

\begin{tcolorbox}[
  colback=blue!6!white,
  colframe=blue!40!black,
  boxrule=0.6pt,
  arc=3pt,
  left=6pt, right=6pt, top=4pt, bottom=4pt
]
  \centering
  \small\textbf{arXiv Preprint} \quad|\quad v1, March 2026
\end{tcolorbox}
\vspace{2mm}

\begin{abstract}
Relative pose estimation is a fundamental task in computer vision with critical applications in SLAM, 
visual localization, and 3D reconstruction. Recently, Relative Pose Regression (RPR) methods have gained 
attention for their end-to-end trainability and inference efficiency. However, existing approaches face 
a fundamental trade-off: traditional
feature-matching pipelines achieve higher accuracy but preclude
end-to-end optimization due to non-differentiable RANSAC, while
recent end-to-end ViT-based regressors enable gradient flow but
demand prohibitive computational resources incompatible with
real-time deployment. We identify the primary bottlenecks in existing RPR approaches as the intrinsic 
coupling between rotation and translation estimation and the lack of effective feature alignment. 
To address these challenges, we propose IUP-Pose, a geometry-driven decoupled iterative framework with 
implicit dense alignment.
Our method first employs a lightweight multi-head bidirectional cross-attention mechanism to implicitly 
align cross-view features. Subsequently, the aligned features undergo a decoupled iterative process 
comprising two rotation prediction stages and one translation prediction stage. To ensure efficiency, 
the two rotation stages share parameters, and the rotation and translation decoders utilize identical 
lightweight architectures. Crucially, we establish a dense information flow between these decoders: 
beyond propagating pose predictions and uncertainty, we explicitly transfer feature maps that are 
iteratively realigned via rotational homography $\mathbf{H}_{\infty}$. 
This geometric guidance ensures that each stage operates on increasingly accurate spatial representations.
Extensive experiments demonstrate that IUP-Pose achieves 73.3\%
$(\text{AUC}@20^{\circ})$ on the MegaDepth1500 benchmark while
maintaining full end-to-end differentiability. Our method uniquely
combines this with extreme efficiency: 70 FPS throughput and only
37M parameters. This establishes a new paradigm for relative pose
estimation that enables real-time deployment on edge devices while maintaining
seamless integration with differentiable 3D perception pipelines.
  \keywords{End-to-End Pose Regression \and
  Geometry-Driven Decoupling \and
  Real-time Efficiency \and Lightweight Architecture}
\end{abstract}

\section{Introduction}
\label{sec:intro}

Relative pose estimation is a cornerstone of computer vision, 
playing a vital role in tasks such as autonomous navigation and 3D reconstruction. 
By directly recovering relative camera motion from image pairs, 
Relative Pose Regression (RPR) enables 3D perception foundation models 
(\eg, VGGT~\cite{wang2025vggt}, MapAnything~\cite{keetha2025mapanything}) to maintain 
spatio-temporally consistent observations of the 3D world. 
Consequently, RPR is increasingly adopted as a versatile sub-task for joint optimization 
to enhance downstream 3D tasks, such as PointMAP generation~\cite{keetha2025mapanything}, 
underscoring its critical role in scalable geometric reasoning.

\begin{figure}[t]
  \centering
  \includegraphics[width=0.8\linewidth]{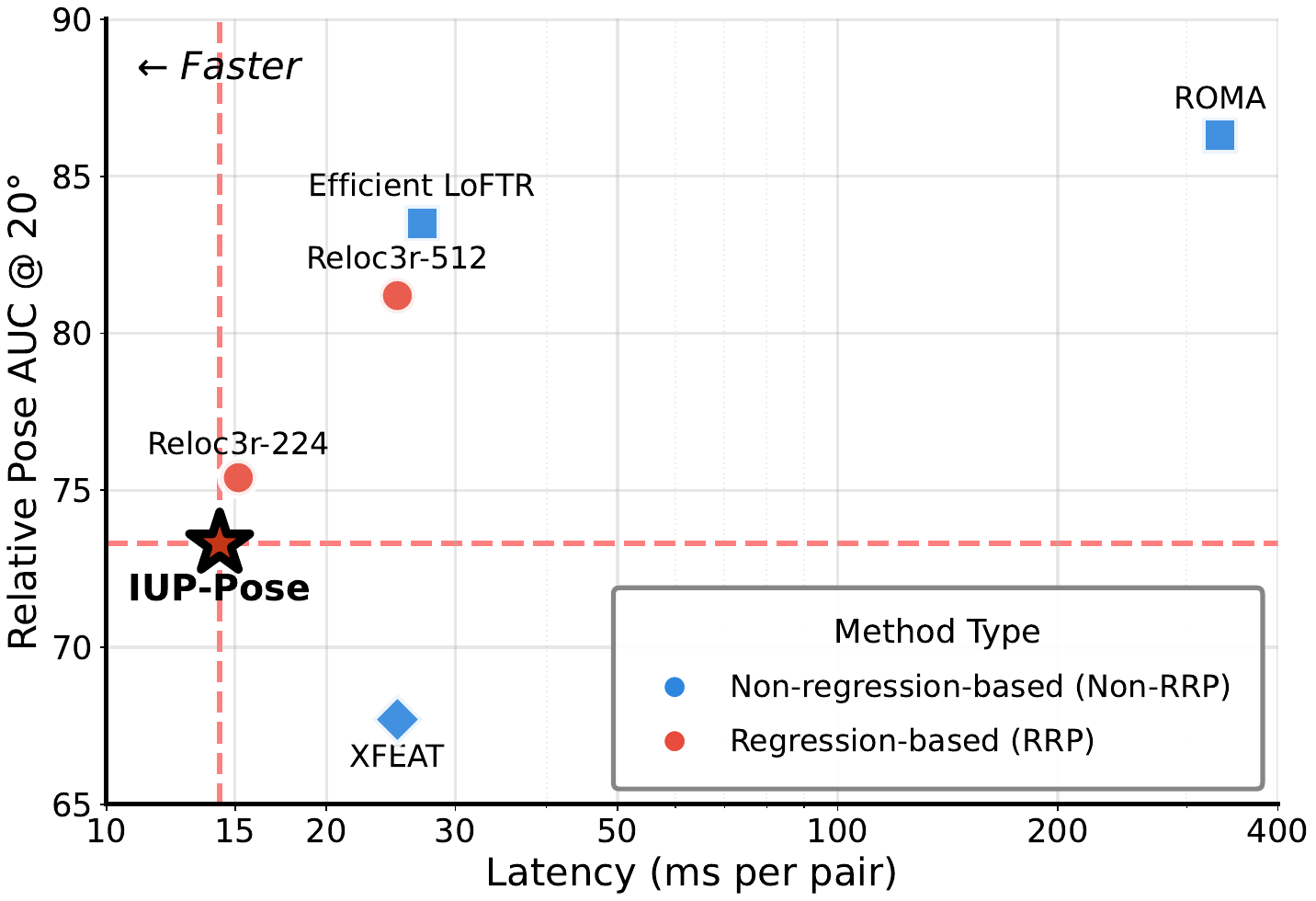}
  \caption{\textbf{Speed-accuracy trade-off on MegaDepth1500.} 
    We evaluate our method against state-of-the-art relative pose estimators. 
    Inference speeds are measured on a single NVIDIA RTX 4090 GPU with 
    latency per image pair (lower is better). 
    Blue markers represent correspondence-based methods; red markers denote 
    regression-based approaches. 
    Our \textbf{IUP-Pose} achieves the lowest latency of 14.3ms (70 FPS), 
    demonstrating superior efficiency while maintaining competitive 
    accuracy at AUC@$20^\circ$ of 73.3\%.}
  \label{fig:performance_comparison}
\end{figure}

Traditional pose estimation methods (non-RPR)~\cite{detone2018superpoint,sarlin2020superglue,sun2021loftr,edstedt2023dkm,
edstedt2024roma,edstedt2025roma,potje2024xfeat,wang2024efficient,lu2024raising} 
follow a two-stage paradigm: 
pixel matching followed by geometric optimization~\cite{hartley2003multiple}.
This workflow establishes correspondences then solves for pose using 
RANSAC~\cite{fischler1981random} with geometric solvers~\cite{nister2004efficient}.
Existing solutions include: 
sparse matching~\cite{lowe2004distinctive,detone2018superpoint,sarlin2020superglue,
mishchuk2017working,mishkin2018repeatability,revaud2019r2d2,tyszkiewicz2020disk,
potje2021extracting,potje2023enhancing,potje2024xfeat,lindenberger2023lightglue} 
(efficient but fails in textureless regions);
dense matching~\cite{edstedt2024roma,edstedt2025roma,edstedt2023dkm,chen2022aspanformer,leroy2024grounding,wang2024dust3r} 
(robust but computationally expensive); 
and semi-dense schemes~\cite{sun2021loftr,potje2024xfeat,lu2024raising,karpur2024lfm,
wang2024efficient,giang2024learning,wang2024homomatcher} 
(balancing efficiency and precision via coarse-to-fine strategies). 
However, this modular design blocks gradient flow, precluding end-to-end 
optimization and hindering integration with differentiable models. 
While~\cite{roessle2023end2end} attempts differentiable integration, 
it suffers from unstable gradients.

In contrast, RPR methods~\cite{wang2025vggt,keetha2025mapanything,melekhov2017relative,chen2021wide,dong2025reloc3r,
abouelnagadistillpose,arnold2022map,ding2019camnet} 
directly infer poses from image pairs, bypassing explicit matching. 
This end-to-end paradigm mitigates cumulative error with pose-level supervision.
Recent foundation models~\cite{wang2025vggt,keetha2025mapanything,dong2025reloc3r} 
achieve comparable performance to non-RPR methods on ScanNet~\cite{dai2017scannet}.
However, heavyweight ViT-based architectures~\cite{dosovitskiy2020image,oquab2023dinov2} 
impose prohibitive computational costs for real-time edge deployment.

We propose IUP-Pose, a lightweight relative pose network with minimal parameters 
and real-time speed. Unlike ViT-based methods~\cite{wang2025vggt,keetha2025mapanything,dong2025reloc3r}, 
we use a ResNet~\cite{he2016deep} backbone. 
Through geometric analysis, we find rotation and translation require different features; 
coupled estimation prevents global optimality. 
We thus propose a geometry-driven decoupled iterative framework with 
a rotate-then-translate head: 
the rotation head predicts axis-angle and uncertainty, 
enabling geometric pre-alignment via rotational homography $\mathbf{H}_\infty$. 
Refined features with rotation results feed the translation head. 
An implicit dense alignment module integrates spatial pyramid pooling 
with cross-attention for global geometric constraints. 
On MegaDepth~\cite{li2018megadepth}, IUP-Pose achieves competitive accuracy 
at 70~FPS with 37M parameters, even under low overlap.

Our key contributions are as follows:
\begin{itemize}
\item \textbf{Decoupled Iterative Refinement:} We propose a
  geometry-driven scheme decoupling pose~\cite{nister2004efficient} into
  rotation and translation subtasks via $\mathbf{H}_\infty$ and
  uncertainty-guided iteration for continuous self-refinement.
\item \textbf{Implicit Dense Alignment:} We design an implicit dense
  alignment module that synergizes spatial pyramid pooling with an
  optimized cross-attention mechanism. This design effectively captures
  global geometric contexts, significantly improving pose regression
  accuracy without compromising real-time speed.
\item \textbf{Competitive Performance:} Experiments on
  MegaDepth~\cite{li2018megadepth} validate our architecture with competitive
  relative pose estimation results (see \cref{fig:performance_comparison}).
\end{itemize}

\section{Related works}
\noindent\textbf{Decoupled Rotation-Translation Regression:}
Recent RPR methods~\cite{wang2025vggt,keetha2025mapanything,dong2025reloc3r}
focus on architectural scaling and extensive pre-training. While achieving
competitive results, these approaches overlook explicit rotation-translation
decoupling, causing model redundancy and high computational cost.

Traditional geometric methods~\cite{kneip2012finding,muhle2022probabilistic,
xu2024accurate,guan2018visual,ding2020homography,zhong2025minimal} have
explored decoupling for robustness and efficiency. Kneip et
al.~\cite{kneip2012finding} proved rotation can be accurately solved
independently, establishing orthogonality between rotational and translational
components in optical flow. M{\"u}hle et al.~\cite{muhle2022probabilistic}
introduced probabilistic epipolar constraints for uncertain features. Kim et
al.~\cite{kim2018low} identified rotation error as the primary VO drift source,
proposing drift-free rotation via lines and planes followed by de-rotated
translation estimation. These insights show decoupling mitigates mutual
interference between pose components.

Homography-based methods~\cite{guan2018visual,ding2020homography,zhong2025minimal}
offer stability in planar scenarios. Guan et al.~\cite{guan2018visual} exploit
ground plane assumptions in autonomous driving. Zhong et
al.~\cite{zhong2025minimal} solve rotation first then translation via homography
to avoid DLT instability under collinearity. However, planar assumptions limit
generalization to arbitrary 3D scenes.

In learning-based RPR, Zhou et al.~\cite{zhou2022decoupledposenet} employed
separate networks with depth for reprojection refinement. Chen et
al.~\cite{chen2021wide} predicted rotation in $SO(3)$ then performed de-rotation
before translation. These frameworks suffer from gradient fragmentation or fail
to propagate geometric constraints. Our IUP-Pose enables full end-to-end
optimization via rotational homography $\mathbf{H}_\infty$ on compressed
features, bridging rotation and translation without auxiliary depth.

\noindent\textbf{Iterative Uncertainty-Guided Refinement:} 
Uncertainty estimation and iterative optimization play pivotal roles in enhancing 
the stability and robustness of computer vision tasks. Recent studies in head pose
estimation~\cite{cantarini2022hhp} have demonstrated that modeling heteroscedastic 
uncertainty via probabilistic loss functions can significantly improve robustness, 
establishing a high correlation between estimated uncertainty and empirical error. 
Similarly, Shukla et al.~\cite{shukla2022vl4pose} utilizes pose likelihood as an uncertainty 
measure to achieve architecture-agnostic active pose learning with high efficiency and 
minimal parameters. In the domain of human pose estimation, 
Li et al.~\cite{li2023pose} proposed an uncertainty-guided iterative refinement framework, 
employing a second-stage optimization guided by initial results and an Uncertainty-Guided 
Self-Attention module to ensure structural consistency. 
Furthermore, Liu et al.~\cite{liu2024hr} modeled uncertainty as the similarity 
between query images and retrieved features, enabling an adaptive number of iterations 
for NeFeS-like~\cite{chen2024neural} pose refinement. 
Other recent works~\cite{li2025ua,dixon2025predicting} have also integrated uncertainty 
modeling to account for observation noise, thereby enhancing pose estimation accuracy.

Parallel to uncertainty modeling, iterative optimization has seen widespread 
adoption across various geometric tasks, including pose estimation~\cite{nguyen2025gotrack}, 
depth estimation~\cite{lipson2021raft}, sparse reconstruction~\cite{tang2024hisplat}, and 
homography estimation~\cite{nie2021depth}. Drawing inspiration from these advancements, 
our IUP-Pose incorporates a tailored iterative refinement module. 
Specifically, we design a three-stage iterative process—consisting of 
two rotation updates followed by one translation update—where each step utilizes 
homography-based alignment to rectify cross-view features. 
By integrating uncertainty guidance at each step, our framework ensures stable 
and effective pose regression, striking a balance between computational efficiency and
geometric precision.

\noindent\textbf{Implicit Dense Alignment:} 
Matching is a pivotal component in pose estimation networks, as it 
effectively aligns input features and directly dictates the 
performance of downstream tasks. Existing approaches can be broadly 
categorized into explicit and implicit matching.

For explicit matching methods, Sarlin et~al.~\cite{sarlin2020superglue}
proposed SuperGlue, which employs Graph Neural Networks (GNNs) to establish
feature correspondences. Similarly, LightGlue~\cite{lindenberger2023lightglue},
LoFTR~\cite{sun2021loftr}, and MatchFormer~\cite{wang2022matchformer} leverage
the attention mechanism~\cite{vaswani2017attention} to perform feature fusion
and matching, where explicit matching quality directly impacts subsequent pose
estimation accuracy.

Conversely, implicit methods, such as~\cite{li2018deepim,melekhov2017relative},
directly fuse cross-view features using CNNs and achieve implicit alignment via
gradient-based regression. Geometric correspondence-based
approaches~\cite{wang2019densefusion,wu2023geometric} match and fuse
corresponding RGB and point cloud information at the pixel level.
Furthermore, Turkoglu et al.~\cite{turkoglu2021visual} utilizes graph networks
to fuse correlations among sparse point features, followed by direct pose
regression via MLPs. More recently, Transformer-based matchers have been
employed by~\cite{xue2023imp,dong2025reloc3r} to simultaneously handle feature
matching and pose regression.

To enhance the performance of our proposed IUP-Pose while ensuring real-time
capability, we adopt an optimized Transformer-based implicit matching strategy.
This approach aligns and fuses multi-view image features, bolstering the
network's robustness against substantial geometric and semantic disparities
across views. Crucially, to maintain computational efficiency, all matching
operations are conducted on $1/32$ scale feature maps, striking a balance
between overall performance and inference speed.
\begin{figure}[tb]
  \centering
  \includegraphics[width=\linewidth]{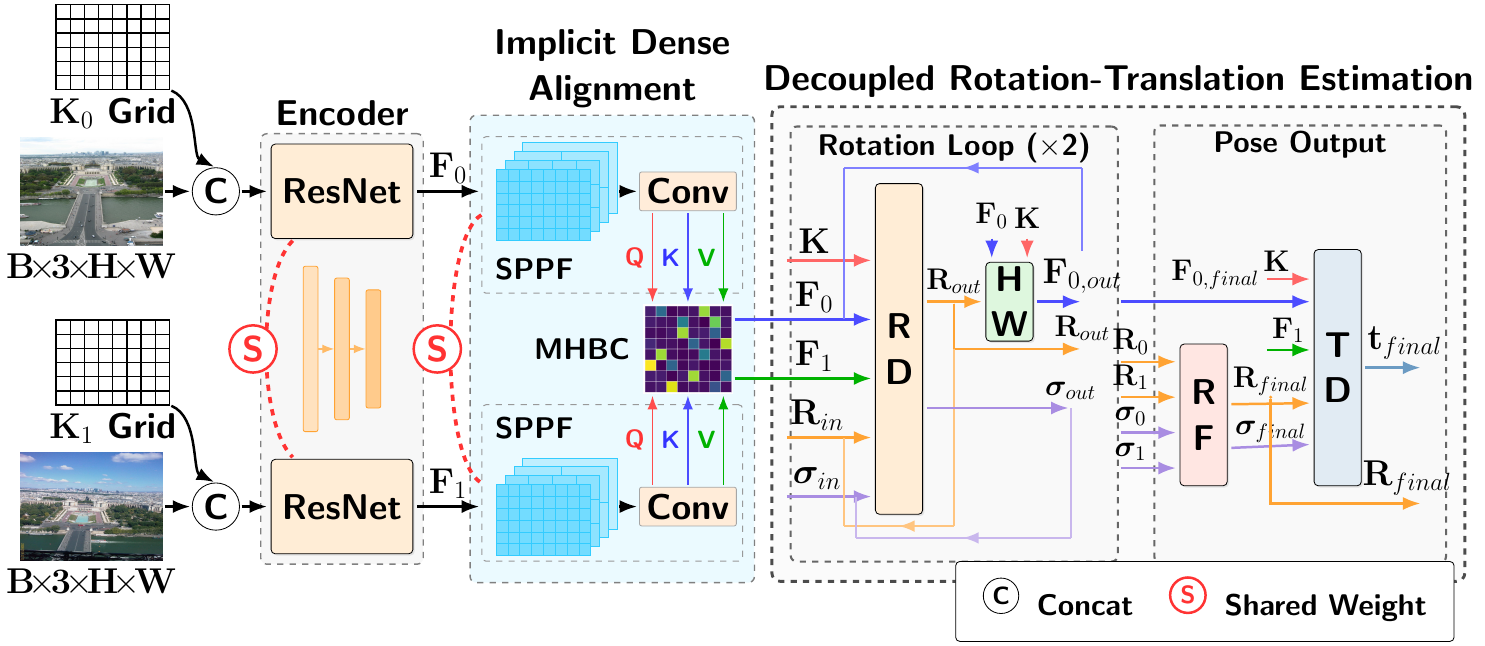}
  \caption{\textbf{Overall architecture of IUP-Pose.} 
  Our framework adopts a decoupled strategy with three main components: 
  \textbf{(1) Input \& Encoder}: RGB images concatenated with normalized 
  coordinates form 5-channel inputs, processed by a ResNet encoder to extract 
  multi-scale features. 
  \textbf{(2) Implicit Dense Alignment (IDA)}: SPPF~\cite{khanam2024yolov5} 
  and multi-head bi-cross attention (\textbf{MHBC}) reduce cross-view domain shift of features. 
  \textbf{(3) Decoupled Rotation-Translation Estimation}: The rotation decoder (\textbf{RD}) 
  iteratively refines $\mathbf{R}$ and $\sigma_R$; homography warp (\textbf{HW}) eliminates 
  rotational disparity between views; rotation fusion (\textbf{RF}) produces final $\mathbf{R}_{final}$; 
  the translation decoder (\textbf{TD}) estimates $\mathbf{t}_{final}$.}
  \label{fig:structure}
\end{figure}
\section{Method}
\subsection{Problem Formulation and Geometric Derivation}
\noindent\textbf{Problem Definition.}
Given a pair of images $\mathbf{I}_0, \mathbf{I}_1$ with overlapping fields of view
and their corresponding intrinsic matrices $\mathbf{K}_0, \mathbf{K}_1$,
the relative pose regression aims to recover the rigid-body transformation
$(\mathbf{R}, \mathbf{t})$ via an end-to-end regression network.
We denote this network as $f_{\boldsymbol{\theta}}(\cdot)$:

\begin{equation}
(\hat{\boldsymbol{\omega}}, \tilde{\mathbf{t}})=f_{\boldsymbol{\theta}}(\mathbf{I}_0,\mathbf{I}_1,\mathbf{K}_0,\mathbf{K}_1),
\quad
\hat{\mathbf{R}}=\mathrm{Exp}(\hat{\boldsymbol{\omega}}^\wedge),
\quad
\hat{\mathbf{t}}=\frac{\tilde{\mathbf{t}}}{\|\tilde{\mathbf{t}}\|_2}.
\end{equation}
Here $\hat{\boldsymbol{\omega}}\in\mathbb{R}^3$ denotes the predicted rotation vector and
${(\cdot)}^\wedge:\mathbb{R}^3\rightarrow \mathfrak{so}(3)$ is the hat operator that maps a vector to a
skew-symmetric matrix in $\mathfrak{so}(3)$. The exponential map $\mathrm{Exp}(\cdot)$ maps
$\mathfrak{so}(3)$ to the rotation group $SO(3)$, ensuring $\hat{\mathbf{R}}\in SO(3)$ 
without explicitly enforcing the orthogonality constraints.
Due to the inherent scale ambiguity in two-view geometry~\cite{hartley2003multiple},
the relative translation can only be recovered up to an unknown positive scale.
Therefore, we represent translation as a unit-norm direction vector on the unit sphere
$\mathbb{S}^2$, i.e., $\|\hat{\mathbf{t}}\|_2=1$, which also improves the generalization of
the translation component~\cite{dong2025reloc3r}.

\noindent\textbf{Geometric Derivation.}
Geometrically, relative pose estimation is a correspondence problem~\cite{hartley2003multiple}.
Under local planarity, the cross-view warp in a neighborhood of pixel $(i,j)$ is well 
approximated by a homography
$\mathbf{H}_{ij}\in\mathbb{R}^{3\times 3}$.
For a 3D plane $\boldsymbol{\pi}=[\mathbf{n}^\top,d]^\top$ observed by two calibrated cameras, the induced homography is
\begin{equation}
\mathbf{H}=\mathbf{K}_2\left(\mathbf{R}+\frac{\mathbf{t}\mathbf{n}^\top}{d}\right)\mathbf{K}_1^{-1},
\end{equation}
where $\mathbf{n}$ is the unit plane normal and $d$ is the signed distance from the first camera center to the plane.
In general scenes, planarity holds only locally. We therefore associate each pixel/feature $(i,j)$ with a tangent plane
parameterized by $\mathbf{n}_{ij}$ and $d_{ij}$, yielding a dense (pixel-varying) homography field
\begin{equation}
\mathbf{H}_{ij}=\mathbf{K}_2\left(\mathbf{R}+\frac{\mathbf{t}\mathbf{n}_{ij}^\top}{d_{ij}}\right)\mathbf{K}_1^{-1}.
\end{equation}
Thus, relative pose estimation can be viewed as modeling a dense homography field and decomposing it into $\mathbf{R}$ and $\mathbf{t}$.

To decouple the global rotational alignment from the local parallax, 
we factor out the rotation matrix $\mathbf{R}$ to the right. 
By inserting the identity matrix $\mathbf{I} = \mathbf{K}_2^{-1}\mathbf{K}_2$, 
we obtain the following multiplicative decomposition:
\begin{equation}
\mathbf{H}_{ij} = \underbrace{\mathbf{K}_2 \left(\mathbf{I} + \frac{\mathbf{t}\mathbf{n}_{ij}^\top}{d_{ij}}\mathbf{R}^{-1}\right)\mathbf{K}_2^{-1}}_{\mathbf{H}_t(i,j)}
\cdot \underbrace{\mathbf{K}_2\mathbf{R} \mathbf{K}_1^{-1}}_{\mathbf{H}_\infty}.
\label{eq:local_homography_decomposition}
\end{equation}
where $\mathbf{H}_t(i,j)$ is the translation-correction term, which is 
highly sensitive to the local geometry (depth and surface normal) of the 
scene, and $\mathbf{H}_\infty$ is the infinite homography (also known as 
the rotational homography), which depends solely on rotation and intrinsics, 
maintaining global spatial consistency across the image. Building on the above 
derivation, IUP-Pose decouples and accurately predicts the rotation 
and translation components.

The overall architecture of IUP-Pose is shown in
\cref{fig:structure}, consisting of three components: Input
Representation and Encoder, Implicit Dense Alignment, and
Decoupled Rotation-Translation Estimation.

\subsection{Input Representation and Encoder}
To mitigate the impact of varying camera intrinsics and establish a 
standardized input representation, we incorporate normalized image plane 
coordinates as an additional spatial prior for the encoder. Specifically, 
for an image $\mathbf{I} \in \{\mathbf{I}_0, \mathbf{I}_1\}$ with its 
corresponding intrinsic matrix $\mathbf{K}$, each pixel 
$\mathbf{p} = [u,v,1]^\top$ is mapped to the canonical image plane via 
the inverse intrinsic transformation:
\begin{equation}
\mathbf{p}_n = [x_n, y_n, 1]^\top = \mathbf{K}^{-1} \mathbf{p}.
\label{eq:normalized_coords}
\end{equation}
We define $\mathbf{C} \in \mathbb{R}^{H \times W \times 2}$ as the 
pixel-wise coordinate map composed of the normalized coordinates 
$(x_n, y_n)$ for all grid locations. The final input to the encoder, 
$\mathbf{X} \in \mathbb{R}^{H \times W \times 5}$, is constructed by 
concatenating the RGB image and the coordinate map along the channel 
dimension:
\begin{equation}
\mathbf{X}_0 = [\mathbf{I}_0 \parallel \mathbf{C}_0], \quad 
\mathbf{X}_1 = [\mathbf{I}_1 \parallel \mathbf{C}_1],
\label{eq:input_concat}
\end{equation}
where $[\cdot \parallel \cdot]$ denotes the concatenation operation. 
This intrinsic-aware design enables the network to implicitly reason 
about the camera's field of view and geometry. 

For efficiency and simplicity, we employ a ResNet~\cite{he2016deep} based encoder 
to extract features. The first convolutional layer is adapted to 
accommodate the 5-channel input tensor. 

\subsection{Implicit Dense Alignment}
Existing matching methods such as XFeat-LighterGlue~\cite{potje2024xfeat} 
achieve good pose estimation accuracy with the keypoint detection mechanism and transformer-based 
feature fusion. To avoid error accumulation caused by explicit keypoint detection and 
realize an end-to-end framework, we adopt an implicit keypoint detection mechanism that achieves local maximum 
perception of features through spatial pyramid pooling, and performs 
feature alignment via a single layer of multi-head bidirectional cross-attention.
Finally, to reduce computational cost and memory consumption, 
only the feature at $1/32$ resolution (stride 32) is fed into the IDA module.

\noindent\textbf{Spatial Pyramid Pooling - Fast (SPPF).}
To efficiently aggregate multi-scale contextual information and implicitly 
detect keypoints, we employ a shared Spatial Pyramid Pooling-Fast (SPPF) module~\cite{khanam2024yolov5} 
for both view features. 
Given an intermediate feature map $\mathbf{F}$, the module first reduces 
its dimensionality via a bottleneck convolution to produce $\mathbf{z}_0$. 
Subsequently, a sequence of max-pooling operations is applied recursively:
\begin{equation}
\mathbf{z}_i = \text{MaxPool}_k(\mathbf{z}_{i-1}), \quad i \in \{1,2,3\},
\label{eq:sppf_pooling}
\end{equation}
where $k$ denotes the kernel size. This recursive structure effectively 
expands the receptive field to multiple scales (equivalent to kernels of 
size $k$, $2k-1$, $3k-2$) and the 
multi-scale features are then concatenated and fused:
\begin{equation}
\mathbf{F}_{sppf} = \text{Conv}_{1\times 1}\left(\text{Concat}(\mathbf{z}_0, 
\mathbf{z}_1, \mathbf{z}_2, \mathbf{z}_3)\right).
\label{eq:sppf_fusion}
\end{equation}
This pooling mechanism not only bolsters the representational capacity but also 
functions as a local maximum perception, or implicit keypoint 
detection technique. Its underlying logic is fundamentally aligned 
with the paradigms of classical keypoint extraction.

\begin{figure}[tb]
  \centering
  \begin{minipage}{0.24\linewidth}
    \centering
    \includegraphics[width=\linewidth,height=0.75\linewidth,keepaspectratio]{2912685234_7c4c21ee78_o.jpg}
    \par\medskip
    (a)
  \end{minipage}\hfill
  \begin{minipage}{0.24\linewidth}
    \centering
    \includegraphics[width=\linewidth,height=0.75\linewidth,keepaspectratio]{3563524378_1c6d70bee9_o.jpg}
    \par\medskip
    (b)
  \end{minipage}\hfill
  \begin{minipage}{0.24\linewidth}
    \centering
    \includegraphics[width=\linewidth,height=0.75\linewidth,keepaspectratio]{4043252998_b660b7dffa_o.jpg}
    \par\medskip
    (c)
  \end{minipage}\hfill
  \begin{minipage}{0.24\linewidth}
    \centering
    \includegraphics[width=\linewidth,height=0.75\linewidth,keepaspectratio]{4142908691_3884438815_o.jpg}
    \par\medskip
    (d)
  \end{minipage}
  \caption{\textbf{Visual disparities in MegaDepth dataset.}
  Representative image pairs from MegaDepth exhibiting significant visual
  challenges: (a) drastic viewpoint changes, (b) inconsistent camera intrinsics,
  (c) illumination differences, and (d) partial occlusions.}
  \label{fig:megadepth_visual_difference}
\end{figure}

\noindent\textbf{Multi-Head Bi-Cross Attention (MHBC).}
In challenging scenarios such as the MegaDepth~\cite{li2018megadepth} dataset, 
image pairs exhibit significant visual disparities due to drastic viewpoint 
changes, illumination variations, structural occlusions and inconsistent camera intrinsics 
as illustrated in \cref{fig:megadepth_visual_difference}. These factors make establishing 
reliable correspondences particularly difficult. To address this challenge, 
we introduce the Multi-Head Bi-Cross Attention (MHBC) module to align corresponding features
and eliminate the cross-view domain shift.
We define a Cross-Attention block with feed-forward network:
\begin{equation}
\mathbf{F}_i' = \text{CrossPSA}(\mathbf{F}_i, \mathbf{F}_j) = \mathbf{F}_i + \Psi(\mathbf{F}_i, \mathbf{F}_j) + \text{FFN}\left(\mathbf{F}_i + \Psi(\mathbf{F}_i, \mathbf{F}_j)\right),
\end{equation}
where $\Psi(\mathbf{F}_i, \mathbf{F}_j)$ computes position-sensitive cross-attention from view $i$ to view $j$:
\begin{equation}
\begin{split}
\Psi(\mathbf{F}_i, \mathbf{F}_j) &= W_p\left[\underbrace{VA^T}_{\text{content}} + \underbrace{W_c A}_{\text{global position}} + \underbrace{\text{PE}(VA^T+W_c A)}_{\text{local position}}\right], \\
A &= \text{softmax}\left(\frac{Q^TK}{\sqrt{d_k}}\right).
\end{split}
\end{equation}
Our design integrates complementary positional information at two levels:
(1) \emph{Global matching receptive field}: The term $W_c A$ encodes the attention distribution itself as positional features, where $W_c \in \mathbb{R}^{d_v \times N_k}$ compresses each query's attention pattern across all $N_k$ key locations into a $d_v$-dimensional feature. This provides global awareness of correspondence patterns.
(2) \emph{Local receptive field}: $\text{PE}(\cdot)$ applies dilated depthwise convolutions with dilation rates $d=2$ and $d=3$ to capture local spatial context in the output feature map.
To reduce computational cost, we use dimension compression with $d_k = 0.5 \times d_v$ for the query and key projections, while maintaining full dimensionality for values.
\begin{figure}[tb]
  \centering
  \begin{minipage}{0.49\textwidth}
    \centering
    \includegraphics[width=\textwidth]{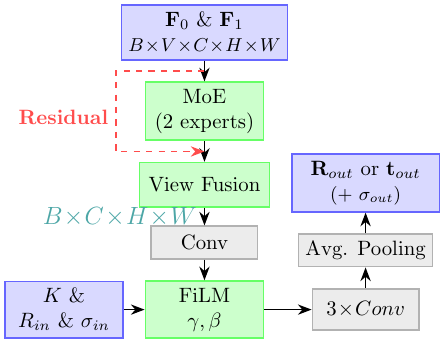}
    \par\medskip
    (a) Overall decoder architecture
  \end{minipage}\hfill
  \begin{minipage}{0.5\textwidth}
    \centering
    \includegraphics[width=\textwidth]{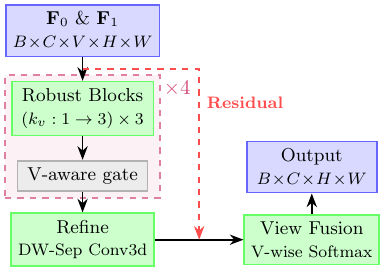}
    \par\medskip
    (b) View Fusion block detail
  \end{minipage}
  \caption{\textbf{Decoder Architecture.}
  Both rotation and translation decoders share this architecture.
  (a) Overall decoder: Multi-view features $\mathbf{F}_0$ and $\mathbf{F}_1$
  pass through MoE adapter (2 experts) and View Fusion to produce fused features
  ($B \!\times\! C \!\times\! H \!\times\! W$). FiLM conditioning with camera
  intrinsics $K$, input pose $R_{in}$, and uncertainty $\sigma_{in}$ is followed
  by convolutional layers and pooling to output $\mathbf{R}_{out}$ (or $\mathbf{t}_{out}$)
  with $\sigma_{out}$. Residual connection (dashed red) preserves input information.
  (b) View Fusion: Features undergo 4 iterations of Robust Bottleneck Blocks with V-aware gates,
  where kernel size $k_v$ grows from 1 to 3 for progressive cross-view mixing.
  After depthwise separable refinement and residual addition (dashed red), V-wise
  softmax attention fuses views to output $B \!\times\! C \!\times\! H \!\times\! W$.}
  \label{fig:pose_decoder}
\end{figure}
\subsection{Decoupled Rotation-Translation Estimation}
The aligned features $F_0$ and $F_1$ are processed through decoupled 
rotation and translation modules. We employ a two-stage rotation 
prediction strategy: the coarse stage predicts an initial rotation 
$\mathbf{R}_{0\to1}^c$ (with axis-angle representation and uncertainty), 
which aligns $F_0$ to $F_1$ via $\mathbf{F}_0^r = \mathbf{H}_\infty^c \cdot \mathbf{F}_0$,
where $\mathbf{H}_\infty^c$ is the rotational homography (see \cref{eq:rot_homo}). 
The refined stage then predicts a residual rotation $\mathbf{R}_{0\to1}^r$ 
conditioned on the coarse prediction and its uncertainty. This 
decomposition extends \cref{eq:local_homography_decomposition} by 
factorizing the infinite homography as 
$\mathbf{H}_\infty = \mathbf{H}_\infty^r \cdot \mathbf{H}_\infty^c$,
where $c$ and $r$ denote the coarse and refined stages, respectively.

The coarse-stage rotation $\mathbf{R}_{0\to1}^c$ is used to align $F_0$ to $F_1$ 
via rotational homography transformation:
\begin{equation}
\label{eq:rot_homo}
\begin{aligned}
\mathbf{H}_\infty^c &= \mathbf{K}_1 \mathbf{R}_{0\to1}^c \mathbf{K}_0^{-1}
\end{aligned}
\end{equation}
where $\mathbf{K}_0$ and $\mathbf{K}_1$ are the camera intrinsics. This alignment 
enables the refined stage to predict a residual rotation $\mathbf{R}_{0\to1}^r$ 
conditioned on the coarse prediction and its uncertainty, which is positively 
correlated with prediction error. Both stages share the same network parameters 
to reduce model complexity.

The final rotation and its uncertainty are obtained by fusing the coarse and 
refined predictions:
\begin{equation}
\begin{aligned}
\mathbf{R}_{0\to1} &= \mathbf{R}_{0\to1}^r \cdot \mathbf{R}_{0\to1}^c, \\
\boldsymbol{\Sigma}_{0\to1} &= \boldsymbol{\Sigma}^r + \mathbf{R}_{0\to1}^r \boldsymbol{\Sigma}^c (\mathbf{R}_{0\to1}^r)^T,
\end{aligned}
\end{equation}
where $\boldsymbol{\Sigma}^c$ and $\boldsymbol{\Sigma}^r$ are the diagonal 
covariance matrices of the coarse and refined uncertainties, respectively. 
The features aligned by the final rotation (with rotational displacements 
eliminated) serve as input for translation prediction, which estimates the 
translation component $\mathbf{H}_t(i,j)$ in \cref{eq:local_homography_decomposition}. 
The translation module employs the similar architecture as the rotation module, 
as illustrated in \cref{fig:pose_decoder}.
\subsection{Loss Function}
We train IUP-Pose with two primary supervision signals: rotation angle and
translation direction. Rotation supervision is applied at both coarse ($c$)
and refined ($r$) stages using geodesic
distance~\cite{huynh2009metrics} and aleatoric
uncertainty~\cite{kendall2017uncertainties} to provide robustness for
rotation outliers, while translation is supervised via normalized direction
error. The overall loss is:
\begin{equation}
\begin{aligned}
  \mathcal{L} &= \mathcal{L}_\text{rot}^c + \mathcal{L}_\text{rot}^r + \mathcal{L}_\text{trans},
\end{aligned}
\end{equation}
For rotation, the network directly predicts axis-angle vectors
$\boldsymbol{\omega}^{c,r}_\text{pred} \in \mathbb{R}^3$ at each stage,
which are then converted to rotation matrices $\mathbf{R}_\text{pred}$ via
the Rodrigues formula. We compute the relative rotation matrix
$\mathbf{R}_\text{rel} = \mathbf{R}^{T}_{\text{pred}}\mathbf{R}_{\text{gt}}$
and supervise via the geodesic distance on SO(3), while rotation uncertainty
follows Laplace negative log-likelihood:
\begin{equation}
\begin{aligned}
  \mathcal{L}_\text{rot}^{c,r} &= \mathcal{L}_\text{angle}^{c,r} + \lambda \mathcal{L}_\text{uncert}^{c,r}, \\
  \mathcal{L}_\text{angle}^{c,r} &= \mathcal{H}_{\delta_r}\left(\arccos\left(\frac{\text{tr}(\mathbf{R}_\text{rel}) - 1}{2}\right)\right), \\
  \mathcal{L}_\text{uncert}^{c,r} &= \sum_{i=1}^{3} \left( |\omega_i^{c,r}| \exp\left(-\tfrac{1}{2}\log\sigma_i^{2}\right) + \tfrac{1}{2}\log\sigma_i^{2} \right),
\end{aligned}
\end{equation}
where $\mathbf{R}_\text{rel} = \mathbf{R}^{T}_{\text{pred}}\mathbf{R}_{\text{gt}}$
is the relative rotation matrix,
$\boldsymbol{\omega}^{c,r} = \log(\mathbf{R}_\text{rel}) \in \mathbb{R}^3$
is the axis-angle error via the SO(3) logarithm map,
$\log\sigma_i^{2}$ is the predicted log-variance for the $i$-th axis at
stage $c$ or $r$, and the uncertainty weight is $\lambda{=}0.1$.
Translation is supervised via the normalized direction error:
\begin{equation}
  \mathcal{L}_\text{trans} = \mathcal{H}_{\delta_t}\left(\arccos\left(\frac{\mathbf{t}_{\text{pred}} \cdot \mathbf{t}_{\text{gt}}}{\|\mathbf{t}_{\text{pred}}\|\|\mathbf{t}_{\text{gt}}\|}\right)\right),
\end{equation}
where $\mathbf{t}_{\text{pred}}$ and $\mathbf{t}_{\text{gt}}$ are the
predicted and ground-truth translation vectors, respectively.
All losses use Huber loss~\cite{huber1964robust} $\mathcal{H}$ with
$\delta_r{=}0.15$~rad and $\delta_t{=}0.5$ to suppress outliers.

\section{Experiments}
\subsection{Experimental Setup}
We train on MegaDepth~\cite{li2018megadepth} and evaluate on MegaDepth1500,
following the data loading and evaluation protocol of
LoFTR~\cite{wang2024efficient}.
Specifically, we use LoFTR's pre-filtered training list consisting of 368
scenes, totaling approximately 1.5 million image pairs.
Input images are resized to $W{=}800$ and $H{=}608$.
For training, we filter image pairs by overlap ratio in $[0.3, 1.0]$,
while the evaluation set remains unfiltered. Similar 
to SRPose~\cite{yin2024srpose}, to further improve performance, 
we also pre-train the model on the ScanNet dataset at scale prior 
to fine-tuning on MegaDepth.

We train for $320{,}000$ optimization steps with a batch size of 20 image
pairs (10 dataloader workers).
We use AdamW~\cite{loshchilov2017decoupled} with learning rate
$2\times10^{-4}$, weight decay $0.01$, and gradient clipping at $5.0$.
We adopt a one-cycle learning-rate schedule with $4{,}000$ warm-up steps
(other settings follow the default configuration).
Training is performed on 4 NVIDIA A800 GPUs.
\subsection{Relative pose estimation}
\noindent\textbf{Evaluation protocol.}
MegaDepth1500 has been widely used in prior
work~\cite{wang2024efficient,lindenberger2023lightglue} and provides test
scenes disjoint from the training set. The test set exhibits challenging
conditions including significant viewpoint changes, illumination variations,
structural changes, and partial occlusions, as shown in
\cref{fig:megadepth_visual_difference}.
Our end-to-end model directly predicts poses without requiring complex
RANSAC-based~\cite{fischler1981random} iterative optimization. During
evaluation, images are kept at the same resolution as training
($W{=}800$, $H{=}608$).

\noindent\textbf{Metrics.}
We report the area under the curve (AUC) of pose error at different thresholds of $5^{\circ}$, $10^{\circ}$, and $20^{\circ}$.
We also measure inference speed in frames per second (FPS) on an NVIDIA RTX 4090 GPU.

\noindent\textbf{Results.}
\Cref{tab:main_results} compares IUP-Pose against state-of-the-art methods.
We categorize competing approaches into two groups: Non-RPR methods that
combine learned matchers with RANSAC-based pose estimation, and RPR approaches
that directly regress camera poses.

\begin{table}[t]
\centering
\caption{Comparison with state-of-the-art methods on MegaDepth1500. 
We report AUC of pose error at different thresholds and inference speed (FPS).}
\label{tab:main_results}
\begin{tabular}{l@{\hspace{8pt}}r@{\hspace{8pt}}r@{\hspace{8pt}}r@{\hspace{8pt}}r}
\toprule
Method & AUC@5$^{\circ}$ & AUC@10$^{\circ}$ & AUC@20$^{\circ}$ & FPS \\
\midrule
\multicolumn{5}{l}{\textit{Non-RPR Methods}} \\
Efficient LoFTR & 56.4 & 72.2 & 83.5 & 37 \\
ROMA & 62.6 & 76.7 & 86.3 & 3 \\
XFEAT & 42.6 & 56.4 & 67.7 & 40 \\
\midrule
\multicolumn{5}{l}{\textit{RPR Methods}} \\
Map-free (Regress-SN) & - & - & <10 & 100 \\
ExReNet (SUNCG) & - & - & <10 & 99 \\
Reloc3r-224 & 39.9 & 59.7 & 75.4 & 66 \\
Reloc3r-512 & 49.6 & 67.9 & 81.2 & 40 \\
IUP-Pose (Ours) & 27.9 & 52.6 & 73.3 & 70 \\
\bottomrule
\end{tabular}
\end{table}

IUP-Pose achieves competitive accuracy while delivering real-time
performance at 70~FPS on an NVIDIA RTX 4090 GPU. Among RPR methods
with comparable accuracy, our model offers the best
speed-accuracy trade-off.
This efficiency advantage stems from our lightweight ResNet backbone combined
with the geometry-driven decoupling strategy, which avoids the computational
overhead of large Vision Transformer architectures used in recent RPR methods.
Unlike traditional feature-matching pipelines that require iterative RANSAC
optimization, our fully differentiable architecture enables end-to-end
training and direct pose regression, eliminating the need for post-processing.

The results demonstrate that our approach strikes a favorable balance between
accuracy and efficiency, making it particularly suitable for real-time
applications such as augmented reality, autonomous navigation, and robotics.
With only 37M parameters, IUP-Pose is significantly more compact than
ViT-based alternatives, enabling deployment on resource-constrained devices.

\subsection{Ablation Study}
To validate the effectiveness of each component in our architecture, we
conduct comprehensive ablation experiments on MegaDepth1500.
Starting from a baseline model, we progressively add key modules and report
their impact on accuracy. \Cref{tab:ablation} summarizes the results.

\begin{table}[t]
\centering
\caption{Ablation study on MegaDepth1500. 
We progressively add components to validate their contributions.
RT-Dec: Rotation-Translation Decoupling; Iter: Iterative Refinement; 
IDA: Implicit Dense Alignment; Uncert: Uncertainty Propagation; 
Homo: Homography Warp; Pre: Pre-training on ScanNet.}
\label{tab:ablation}
\begin{tabular}{cccccc@{\hspace{8pt}}r@{\hspace{8pt}}r@{\hspace{8pt}}r}
\toprule
RT-Dec & Iter & IDA & Uncert & Homo & Pre & AUC@5$^{\circ}$ & AUC@10$^{\circ}$ & AUC@20$^{\circ}$ \\
\midrule
        &     &     &        &      &      & 4.1 & 14.2 & 31.3  \\
\checkmark &     &     &        &      &      & 8.4 & 19.0 & 37.9  \\
\checkmark & \checkmark &     &        &      &      & 10.6 & 23.3 & 42.9  \\
\checkmark & \checkmark & \checkmark &        &      &      & 14.8 & 28.5 & 45.2  \\
\checkmark & \checkmark & \checkmark & \checkmark &      &      & 16.8 & 31.3 & 48.8  \\
\checkmark & \checkmark & \checkmark & \checkmark & \checkmark &      & 22.5 & 40.9 & 57.6  \\
\checkmark & \checkmark & \checkmark & \checkmark & \checkmark & \checkmark & 27.9 & 52.6 & 73.3  \\
\bottomrule
\end{tabular}
\end{table}

\noindent\textbf{RT-Dec} brings the first substantial gain
(AUC@$5^\circ$: $4.1\!\to\!8.4$), confirming that decoupling
rotation and translation reduces mutual interference.
\textbf{Iter} further improves via coarse-to-fine residual
refinement ($8.4\!\to\!10.6$).
\textbf{IDA} (SPPF+MHBC) yields the largest gain at strict
thresholds ($10.6\!\to\!14.8$ at AUC@$5^\circ$), validating
that implicit cross-view alignment significantly improves
correspondence quality.
\textbf{Uncert} provides consistent improvement across all
thresholds ($+2.0$/$+2.8$/$+3.6$) by guiding the decoder
toward reliable regions.
\textbf{Homo} delivers the largest single inference-time boost
($+5.7$/$+9.6$/$+8.8$) by eliminating rotational disparity
before translation estimation.
Finally, \textbf{Pre}-training on ScanNet~\cite{dai2017scannet}
contributes the most at the relaxed threshold
($+5.4$/$+11.7$/$+15.7$), indicating strong generalization
from the indoor-scene prior.

\subsection{Qualitative Results}
\begin{figure}[t]
  \centering
  \includegraphics[width=0.95\linewidth]{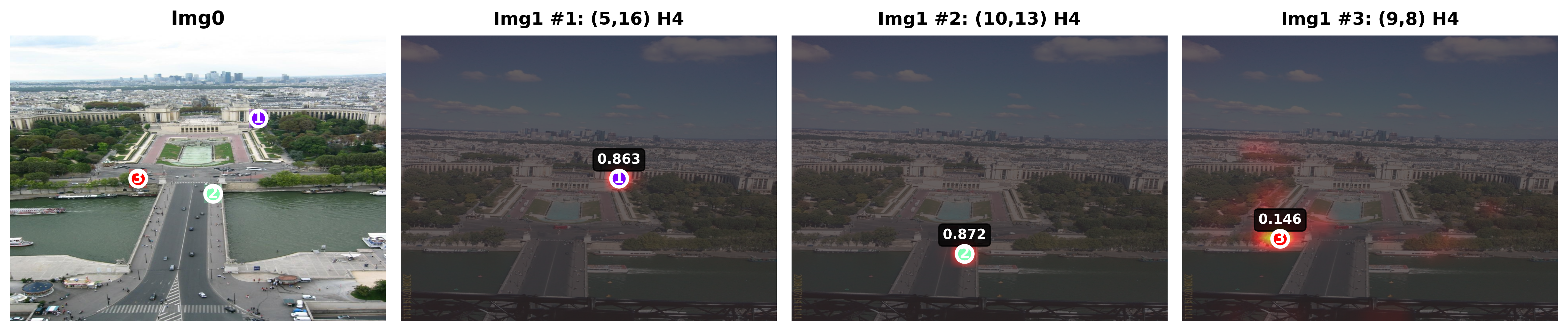}
  \vspace{-2mm}
  \caption*{(a) IDA attention heatmaps}
  \vspace{3mm}
  
  \includegraphics[width=0.95\linewidth]{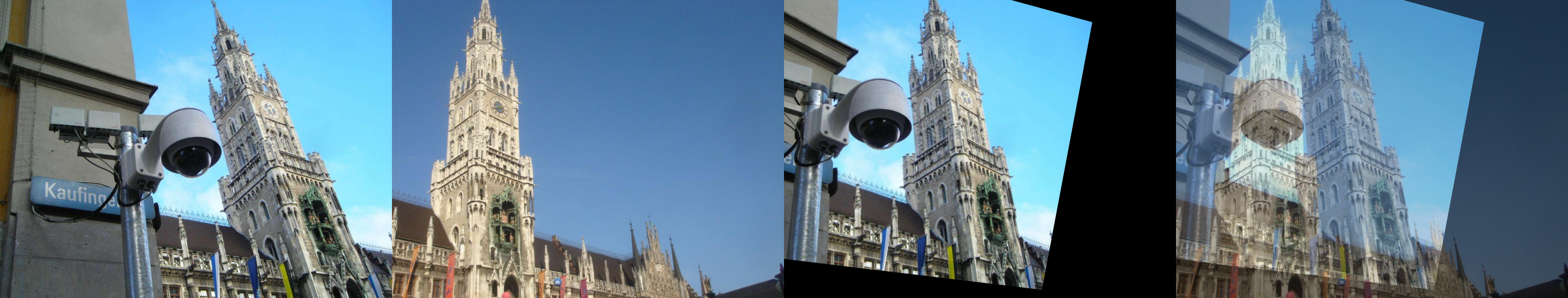}
  \vspace{-2mm}
  \caption*{(b) Homography warping effect}
  
  \caption{\textbf{Qualitative analysis.} 
    (a) Cross-view attention heatmaps from MHBC: query points in Img0 
    (left, colored circles) attend to semantically corresponding regions 
    in Img1 (right panels show attention scores). 
    (b) Feature alignment via rotational homography $\mathbf{H}_\infty$: 
    left shows misaligned features before warping, right shows aligned 
    features after applying $\mathbf{H}_\infty$ to eliminate rotational 
    disparity.}
  \label{fig:qualitative}
\end{figure}

\Cref{fig:qualitative} visualizes the learned correspondences via IDA 
attention heatmaps and the effect of homography-based warping.
Without explicit matching supervision, the MHBC mechanism successfully 
learns semantically meaningful cross-view correspondences: high-attention 
regions (scores $0.863$/$0.872$) accurately localize matching structures 
across viewpoints, while low-attention regions ($0.146$) correspond to 
textureless or occluded areas, demonstrating uncertainty-aware feature 
selection.
The homography warping visualization (b) shows how $\mathbf{H}_\infty$ 
eliminates rotational disparity, aligning features before translation 
estimation to reduce geometric ambiguity.

\subsection{Performance Across Overlap Ratios}

\noindent
\begin{minipage}[t]{0.38\linewidth}
  \centering
  \captionof{table}{AUC@10$^{\circ}$ on MegaDepth1500
    across overlap ranges.}
  \label{tab:overlap}
  \vspace{2pt}
  \begin{tabular}{l@{\hspace{6pt}}r}
    \toprule
    Overlap Range & IUP-Pose \\
    \midrule
    {[0.0, 0.1]} & 24.8 \\
    {[0.1, 0.4]} & 50.5 \\
    {[0.4, 0.7]} & 57.5 \\
    {[0.7, 1.0]} & 58.3 \\
    \midrule
    Overall       & 52.6 \\
    \bottomrule
  \end{tabular}
\end{minipage}%
\hfill
\begin{minipage}[t]{0.59\linewidth}
  \small
  To evaluate robustness under varying viewing conditions,
  we analyze performance across different overlap ratios
  (\cref{tab:overlap}).
  In low-overlap scenarios ({[0.0, 0.1]}), where traditional
  feature matching often fails due to insufficient
  correspondences, our end-to-end architecture leverages the
  learned implicit alignment through the IDA module.
  The geometry-driven decoupling strategy is particularly
  beneficial here, exploiting the natural structure of rigid
  transformations even when explicit correspondences are sparse.
  For high-overlap scenarios ({[0.4, 1.0]}), IUP-Pose achieves
  competitive accuracy while maintaining its efficiency
  advantage, validating robustness across the full overlap
  spectrum.
\end{minipage}

\section{Conclusion}
We propose IUP-Pose, a geometry-driven decoupled iterative framework 
for real-time relative pose estimation.
By decomposing the 5-DoF pose regression into independent rotation and 
translation subproblems, our method exploits the natural structure of 
rigid transformations through rotational homography $\mathbf{H}_\infty$ 
and uncertainty-guided refinement.
The implicit dense alignment module establishes cross-view correspondences 
without explicit matching supervision, achieving competitive accuracy on 
MegaDepth1500 (AUC@10$^\circ$: 52.6\%) while maintaining real-time 
throughput at 70~FPS with only 37M parameters.

We observe that performance degrades on indoor datasets such as ScanNet, 
where large rotations and dominant translations cause homography warping 
to project corresponding pixels outside the image bounds.
While the IDA module partially mitigates this issue through implicit 
alignment, future work could explore adaptive warping strategies or 
multi-scale feature pyramids to better handle extreme viewpoint changes 
in confined spaces.

%
%
\bibliographystyle{splncs04}
\bibliography{main}
\end{document}